# An Image Processing Pipeline for Autonomous Deep-Space Optical Navigation *

Eleonora Andreis [†], Paolo Panicucci [‡], and Francesco Topputo [§]
*Politecnico di Milano, 20156, Milan, Italy*

A new era of space exploration and exploitation is fast approaching. A multitude of spacecraft will flow in the future decades under the propulsive momentum of the new space economy. Yet, the flourishing proliferation of deep-space assets will make it unsustainable to pilot them from ground with standard radiometric tracking. The adoption of autonomous navigation alternatives is crucial to overcoming these limitations. Among these, optical navigation is an affordable and fully ground-independent approach. Probes can triangulate their position by observing visible beacons, e.g., planets or asteroids, by acquiring their line-of-sight in deep space. To do so, developing efficient and robust image processing algorithms providing information to navigation filters is a necessary action. This paper proposes an innovative pipeline for unresolved beacon recognition and line-of-sight extraction from images for autonomous interplanetary navigation. The developed algorithm exploits the k-vector method for the non-stellar object identification and statistical likelihood to detect whether any beacon projection is visible in the image. Statistical results show that the accuracy in detecting the planet position projection is independent of the spacecraft position uncertainty. Whereas, the planet detection success rate is higher than 95% when the spacecraft position is known with a $3\sigma$ accuracy up to $10^5$ km.

## Nomenclature

| | | |
|---|---|---|
| $\mathcal{N}$ | = | inertial reference frame defined as $\mathcal{N} = \{n, \boldsymbol{n}_1, \boldsymbol{n}_2, \boldsymbol{n}_3\}$ |
| $^{\mathcal{N}}_{h}\boldsymbol{r}_{\text{bc}}$ | = | beacon position vector in homogeneous coordinate in $\mathcal{N}$ |
| $^{\mathcal{N}}\boldsymbol{r}_{\text{bc}}$ | = | beacon position vector in $\mathcal{N}$ |
| $^{\mathcal{N}}\boldsymbol{r}$ | = | spacecraft position vector in $\mathcal{N}$ |
| $^{\mathcal{N}}\boldsymbol{\rho}$ | = | line-of-sight direction of the beacon as seen by the spacecraft in $\mathcal{N}$ |
| $C$ | = | 3D camera reference frame defined as $C = \{c, \boldsymbol{c}_1, \boldsymbol{c}_2, \boldsymbol{c}_3\}$ |
| $^{C}\boldsymbol{\rho}$ | = | line-of-sight direction of the beacon as seen by the spacecraft in $C$ |

---

| | | |
|---|---|---|
| $[CN]$ | = | attitude matrix from $\mathcal{N}$ to $\mathcal{C}$ |
| $[R_i]$ | = | rotation matrix around the $i$th axis |
| $\alpha$ | = | right ascension angle of the camera |
| $\delta$ | = | declination angle of the camera |
| $\phi$ | = | twist angle of the camera |
| $\mathbb{C}$ | = | 2D camera reference frame defined as $\mathbb{C} = \{C, \boldsymbol{C}_1, \boldsymbol{C}_2\}$ |
| $^{\mathbb{C}}\boldsymbol{R}_{\text{bc}}$ | = | planet position projection vector in $\mathbb{C}$ |
| $^{\mathbb{C}}_{h}\boldsymbol{R}_{\text{bc}}$ | = | planet position projection vector in homogeneous coordinates in $\mathbb{C}$ |
| $[K]$ | = | intrinsic camera matrix |
| $f$ | = | camera focal length |
| $I_{\text{thr}}$ | = | threshold value expressed in pixel intensity for removing the background noise |
| $\mu_I$ | = | mean intensity over the image |
| $\sigma_I$ | = | intensity standard deviation over the image |
| $T$ | = | tuning parameter for the dynamic thresholding |
| $I_{i,j}$ | = | intensity of the pixel $(i, j)$ |
| $(X_{i,j}, Y_{i,j})$ | = | coordinates of the pixel $(i, j)$ |
| $w_{i,j}$ | = | weighting parameter of the pixel $(i, j)$ |
| $(X_c, Y_c)$ | = | centroid coordinates |
| $I_{00}, I_{01}, I_{10}$ | = | image momenta |
| $\gamma$ | = | interstellar angle |
| $\boldsymbol{K}$ | = | k-vector available onboard |
| $\boldsymbol{I}, \boldsymbol{J}$ | = | vectors available onboard where the star pairs IDs are stored |
| $[^{\mathcal{N}}s]$ | = | matrix whose columns contain the stars line-of-sight directions in $\mathcal{N}$ |
| $[^{C}s]$ | = | matrix whose columns contain the stars line-of-sight directions in $C$ |
| $n_R$ | = | number of samples for the application of the RANSAC algorithm |
| $\boldsymbol{s}_i$ | = | $i$th group of three stars in the image |
| $\boldsymbol{e}$ | = | spacecraft rotation principal axis |
| $\theta$ | = | spacecraft rotation principle angle |
| $t$ | = | threshold angle in the RANSAC algorithm |
| $^{\mathbb{C}}\boldsymbol{R}_{\text{bc}_0}$ | = | beacon expected position projection |
| $[P]$ | = | covariance matrix of the beacon position projection |
| $[F]$ | = | Jacobian matrix of the mapping linking $^{\mathbb{C}}\boldsymbol{R}_{\text{bc}}$ with the spacecraft pose and beacon position |
| $[S]$ | = | uncertainty covariance matrix of the probe pose and beacon position |
| $\boldsymbol{q}$ | = | spacecraft rotation quaternion defined as $\boldsymbol{q} = (q_0, \boldsymbol{q}_v)^\top$ |
| $[A]$ | = | estimated attitude matrix of the probe |
| $[\mathbb{I}_n]$ | = | identity matrix of dimension $n$ |
| $[\mathbb{0}_{n \times m}]$ | = | zero $n \times m$ matrix |
| $[\boldsymbol{x}^\wedge]$ | = | skew symmetric matrix associated with the cross product $\boldsymbol{x} \times \boldsymbol{y}$. |
| $\sigma_i$ | = | standard deviation of the element $i$ |
| a | = | $3\sigma$ covariance ellipse semimajor axis |
| b | = | $3\sigma$ covariance ellipse semiminor axis |
| $\psi$ | = | $3\sigma$ covariance ellipse orientation |
| $F$ | = | f-number of the camera |
| $Q_e \times T_{\text{lens}}$ | = | quantum efficiency × lens transmission of the camera |
| $\sigma_d$ | = | defocus level of the camera |
| $m_{\text{lim}}$ | = | apparent magnitude threshold considered for the creation of the onboard catalogs |
| $\epsilon$ | = | k-vector range error |
| $\boldsymbol{\mu}_{\text{err}}$ | = | vector representing the mean of the planet position projection errors |
| $[P_{\text{err}}]$ | = | matrix representing the covariance of the planet position projection errors |
| $\sigma_{\text{ErrRot}}$ | = | standard deviation of the rotation error |

## I. Introduction

A new era of deep-space exploration and exploitation is fast approaching. In the next decade, a flourishing growth of probes will be launched in interplanetary space under the propulsive momentum of the new space economy. Nowadays, deep-space probes are mostly piloted with standard radiometric tracking. Yet, at the current pace, since this approach heavily relies on limited resources, such as ground stations and dedicated teams, its adoption will become unsustainable soon. In other terms, the exploitation of radiometric tracking will hamper the proliferation of deep-space assets [2]. Self-driving deep-space probes, which are free from ground-based support, would overcome these limitations [3]. From a navigation perspective, spacecraft can determine their state by observing the external environment with cameras; major [4] and minor bodies [5] observations can be exploited to triangulate the spacecraft position, provided their ephemerides are known [6]. Yet, in deep space, planets and asteroids are unresolved and their light falls in one pixel only of the observing camera. Thus, they can not be distinguished at first sight from the stars. Indeed, one of the most relevant issues of far-range vision-based navigation (VBN) consists of the recognition and labeling of the celestial beacons in the image against the stellar background.

In 1998, the Deep-Space 1 (DS1) mission [7, 8] proved the feasibility of estimating the probe state in deep space by observing visible asteroids in the asteroid belt [9]. The basic concept of the DS1 onboard autonomous navigation system was to feed an orbit determination algorithm with the unresolved targets' inertial Line-Of-Sight (LoS) vectors extracted from the images taken during the cruise phase. Following studies were then performed in Broschart et al. [5], which shows the achievable positioning accuracies resulting from the exploitation of visible asteroids as beacons between the orbits of Mercury and Jupiter. Yet, when cameras of limited performances are adopted onboard low-priced miniaturized probes, such as CubeSats, asteroids can not be visible in the sensor frame. Only brighter celestial bodies, like planets, can be observed for far-range optical navigation. In this context, the EXTREMA (Engineering Extremely Rare Events in Astrodynamics for Deep-Space Missions in Autonomy) project [10], awarded an ERC Consolidator Grant in 2019, aims to understand how to enable deep-space, limited-budget spacecraft to perform navigation, guidance, and control operations in complete autonomy with respect to ground.

Previous works focus on the implementation of onboard orbit determination algorithms to estimate the probe state [4, 6, 11]. In addition, while detailed pipelines for Image Processing (IP) at mid- and close-range is available [12–14], the case of deep space is still an open issue.

This work contributes to the state-of-the-art by developing an innovative IP pipeline for the extraction of the beacon projection from camera image in the context of deep-space autonomous navigation. The procedure is composed of two parts. First, the lost-in-space attitude determination problem is solved by adopting the k-vector method [15], which also allows the non-stellar objects recognition in the image. Second, the beacon identification is performed by evaluating the beacon position projection uncertainty ellipse and by selecting the non-stellar object, if any, contained in it. The IP pipeline described is then applied to deep-space CubeSats in the framework of the EXTREMA project.

The paper is structured as follows. First, Sec. II summarizes the necessary mathematical preliminaries. Then, in Sec. III the methodology followed for the development of the IP pipeline is defined. In addition, Sec. IV presents a general study of the algorithm behavior by focusing on the off-nominal scenarios during the extraction of the beacon position projection. Finally, Sec. V gathers the results obtained by the application of the proposed IP pipeline.

## II. Projective Geometry Preliminaries

Let ${}^N\boldsymbol{r}_{\text{bc}}$ and ${}^N_h\boldsymbol{r}_{\text{bc}}$ be the beacon position vector in the inertial reference frame $\mathcal{N} = \{n, \boldsymbol{n}_1, \boldsymbol{n}_2, \boldsymbol{n}_3\}$ expressed with non-homogeneous and homogeneous coordinates, respectively, and let ${}^N\boldsymbol{r}$ be the spacecraft position in non-homogeneous coordinates. Interested readers can refer to [16] for further details about homogeneous vector representation. With reference to Fig. 1a, the position of the beacon as seen by the spacecraft in $\mathcal{N}$ is described as

$$ {}^N\boldsymbol{\rho} = {}^N\boldsymbol{r}_{\text{bc}} - {}^N\boldsymbol{r} \tag{1} $$

Let a projective camera observe the beacon. The camera frame is defined as $C = \{c, c_1, c_2, c_3\}$. The vector ${}^N\rho$ can be transformed in $C$ though a passive rotation from $N$ to $C$ and applied by the attitude matrix $[CN]$:

$$^C\rho = [CN]\,{}^N\rho \tag{2}$$

For the definition of the attitude matrix, the Axis-Azimuth representation is adopted [17]. By assuming that the camera boresight is coincident with the third axis of the spacecraft-fixed reference frame,

$$[CN] = [R_3(\alpha)]\,[R_2(\pi/2 - \delta)]\,[R_3(\phi)] \tag{3}$$

where $[CN]$ is obtained through a succession of counterclockwise rotations taking into account the camera pointing angles: right ascension $\alpha \in [0°, 360°]$, declination $\delta \in [-90°, 90°]$, and twist angle $\phi \in [0°, 360°]$ [18].
Once ${}^C\rho$ is computed (see Eq. (2)), the 3D point is projected on the image plane by exploiting the pinhole camera model.

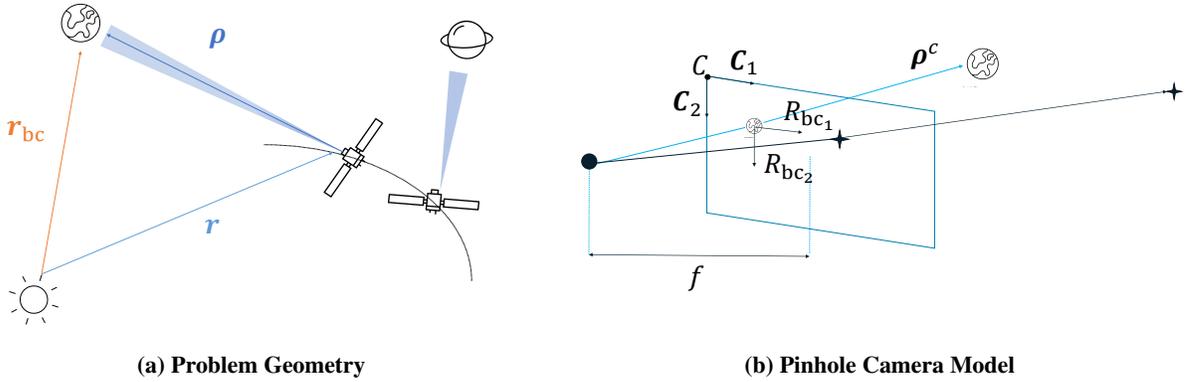

(a) Problem Geometry    (b) Pinhole Camera Model

Fig. 1  Problem and Projective Geometry Preliminaries

With reference to Fig. 1b, let $\mathbb{C} = \{C, C_1, C_2\}$ be the 2D camera reference frame where $C_1$ points to the right, $C_2$ downward, and the reference frame center $C$ is placed at the upper left-hand corner of the image. The projection of the planet position in $\mathbb{C}$ is ${}^\mathbb{C}R_{bc}$ and the transformation of the beacon position vector in homogeneous coordinates from $N$ to $\mathbb{C}$ is compactly described as

$$^\mathbb{C}_h R_{bc} = [K]\,\underbrace{[CN]\begin{bmatrix}\mathbb{I}_3 & -{}^N r\end{bmatrix}{}^N_h r_{bc}}_{^C\rho} \tag{4}$$

where $[K]$ is the intrinsic camera matrix [19]. Eventually, the beacon position projection in $\mathbb{C}$ in non-homogeneous coordinates ($^{\mathbb{C}}\boldsymbol{R}_{\text{bc}}$) becomes

$$^{\mathbb{C}}\boldsymbol{R}_{\text{bc}} = \begin{pmatrix} ^{\mathbb{C}}R_{\text{bc}_1} \\ ^{\mathbb{C}}R_{\text{bc}_2} \end{pmatrix} = \begin{pmatrix} \frac{^{\mathbb{C}}_h R_{\text{bc}_1}}{^{\mathbb{C}}_h R_{\text{bc}_3}} \\ \frac{^{\mathbb{C}}_h R_{\text{bc}_2}}{^{\mathbb{C}}_h R_{\text{bc}_3}} \end{pmatrix} \quad (5)$$

where $^{\mathbb{C}}_h R_{\text{bc}_i}$ is the $i$th component of the beacon position projection in $\mathbb{C}$ in homogeneous coordinates. Through an analogous procedure, stars are projected on the camera frame by considering that they are laying on the plane at infinity. Stars positions are usually stored in catalogs, such as the Hipparcos' [20], which provides their right ascension and declination on the celestial sphere [18].

## III. Methodology

The final goal of the algorithm is the detection of the beacon in the digital image and the extraction of its position projection. From the latter, the beacon LoS direction, which has been proven to be a valuable observable for deep-space VBN [4, 11], can be straightforwardly retrieved.

To achieve its goal, the IP algorithm performs two steps sequentially:

1) The determination of the spacecraft attitude through star asterism identification.
2) The beacon detection in the image.

The flowchart of the proposed IP pipeline is shown in Figure 2. First, a search-less method based on the k-vector technique [15, 21] and bulked up with a RANSAC [16] procedure is adopted. This provides two results: 1) the match of the stellar objects detected in the deep-space image with the associated identifiers (ID) stored in the star catalog, and 2) the detection of non-stellar objects (spikes) in the image. Once the stars ID are known, the probe orientation is computed by solving the Wahba problem [22] between the stars LoS directions measured in the camera frame and the ones matched in the inertial frame catalog. With the knowledge of the probe attitude and an estimation of its position, a prediction of the beacon position projection in the image can be computed. In addition, its uncertainty can be defined by assuming known the uncertainty of the probe pose, i.e., probe position and orientation, and of the beacon position. As the computed statistical momenta, i.e., the expected beacon projected position and its uncertainty, define the Gaussian probability to find a beacon in that portion of the image, this information can be exploited to identify the celestial body and extract its position projection. Note that if one or more spikes are contained within the beacon projection uncertainty ellipse, the closest spike to the expected beacon projection position is the highest probable to be the desired beacon projection, therefore it is identified as the beacon itself.

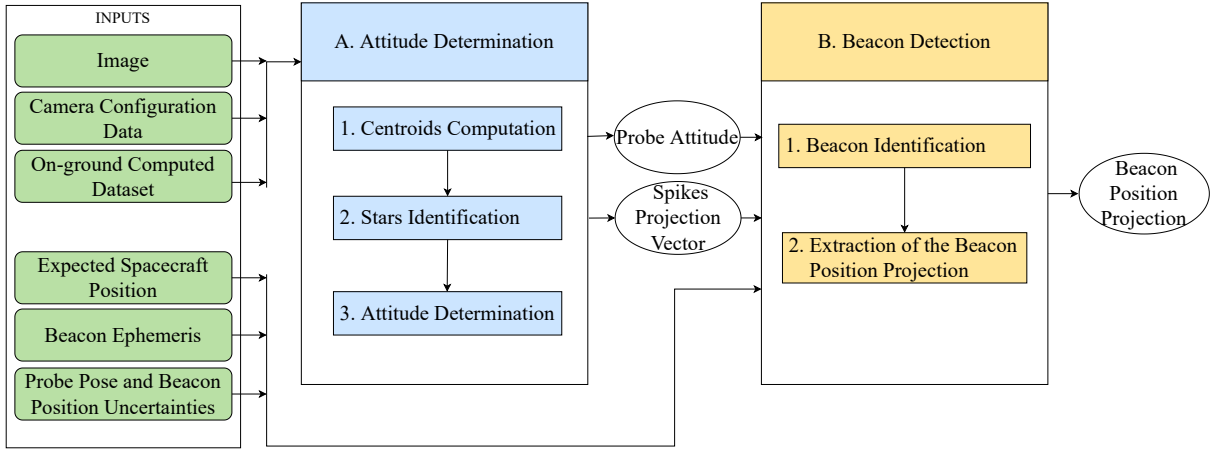

**Fig. 2 Workflow of the proposed IP pipeline.**

The following assumptions represented with the green rounded rectangles in Fig. 2 are made:

1) The camera is previously calibrated, thus the camera intrinsic parameters are known.
2) The catalogs needed for the star identification are available.
3) A rough estimation of the probe position is known.
4) Beacons ephemeris are available onboard.
5) The probe pose and beacons position uncertainties are known.

### A. Attitude Determination

The attitude determination is based on the execution of three steps, namely:

1) The centroids of the bright objects in the image are found.
2) A subset of the centroids in the image are matched with the onboard-stored star catalog, which contains the stars LoS directions in the inertial reference frame.
3) The attitude of the probe is determined by solving the Wahba problem between the stars LoS directions in the camera and inertial reference frame.

*1. Centroids Computation*

The first step of every optical sensor that has to perform star identification consists of the determination of the location of the bright objects in the image. In deep space, the observed objects are mostly unresolved. For cameras focused at infinity, the light from each star or beacon falls in one pixel only. Since the extracted centroid is the center of the pixel itself, this implies that only pixel accuracy can be achieved. A strategy adopted to achieve subpixel precision in centroid extraction from unresolved objects is to intentionally defocus the camera to spread the incoming light over multiple pixels [23]. When a defocused image is acquired, the centroids of the bright pixels are simply found by computing the center of brightness. In this work, the procedure presented [18] in is applied. This one can be subdivided into the following steps:

1) A threshold value $I_{\text{thr}}$, expressed in pixel intensity, is set up to remove the background noise. This is of paramount

importance to select those pixels to consider in the computation. The value of $I_{\text{thr}}$ is determined by applying a dynamic thresholding method that can be tuned to improve the performance of the algorithm:

$$I_{\text{thr}} = \mu_I + T\sigma_I \qquad (6)$$

where $\mu_I$ is the intensity mean over the image, $\sigma_I$ is the intensity standard deviation over the image, and $T$ is the tuning parameter.

2) By thresholding the image using $I_{\text{thr}}$, pixels brighter than the threshold are identified. They form connected portions of the image which can be delimited with squared Region Of Interest (ROI) with a margin of one pixel on each side. All the pixels inside one ROI define a single stellar or non-stellar object projection. Thus, the centroid of the object can be computed by using the pixels inside the associated ROI.

3) The image momenta $I_{10}$, $I_{01}$, and $I_{00}$ inside each ROI are found as

$$I_{00} = \sum_{(i,j)\in \text{ROI}} I_{i,j}\, w_{i,j} \qquad I_{10} = \sum_{(i,j)\in \text{ROI}} X_{i,j}\, I_{i,j}\, w_{i,j} \qquad I_{01} = \sum_{(i,j)\in \text{ROI}} Y_{i,j}\, I_{i,j}\, w_{i,j} \qquad (7)$$

where $X_{i,j}$ and $Y_{i,j}$ are the pixel coordinates, $I_{i,j}$ its intensity, and $w_{i,j}$ the weighting parameter associated with the pixel $(i, j)$. In this work, the weighting parameter is defined as $w_{i,j} = \frac{I_{i,j}}{I_{i,j_{max}}}$ to give more importance to brighter pixels inside the ROI [24].

4) Once the image momenta for a ROI are computed, the sub-pixel centroid coordinates associated with that ROI are found as

$$X_c = \frac{I_{10}}{I_{00}} \qquad Y_c = \frac{I_{01}}{I_{00}} \qquad (8)$$

*2. Stars Identification*

The goal of the star identification procedure is to recognize, among the found centroids, the ones that are stars projection in the image. To do so, the problem is rewritten as a registration problem whose goal is to find the correct matching between the observed star asterism and the cataloged stars in the inertial frame. For this purpose, the search-less algorithm (SLA) introduced in [15] is adopted. In this work, the SLA has been preferred over the binary search technique [25] for its higher speed gain rate (from 10 to more than 50 times [26]) and for its capability to identify spikes among the bright objects in the image. This latter is maybe the most important feature since the navigation beacons are just searched among these spikes. Finally, an additional interesting characteristic of the SLA is its independence from any a priori attitude guess and magnitude information.

To be adopted, the SLA requires computing on ground a vector of integers, the k-vector, which contains information for the stars matching starting from the chosen stars invariant. In this work, the invariant chosen to build the star catalog is the interstellar angle $\gamma$ which is defined as

$$\gamma_{ml} = \arccos({}^{N}\boldsymbol{\rho}_m^\top {}^{N}\boldsymbol{\rho}_l) \tag{9}$$

where ${}^{N}\boldsymbol{\rho}_m$ and ${}^{N}\boldsymbol{\rho}_l$ are the LoS directions to the $m$th and $l$th star, respectively. Note that the interstellar angle is an invariant parameter with respect to the reference frame considered. Indeed, when the onboard catalog is derived, the interstellar angles are computed from the stars LoS directions expressed in the $N$ frame. However, in the operative phase, these angles are matched with the observed interstellar angles obtained from the stars LoS direction in the $C$ frame. The k-vector on-ground computation requires several steps described hereunder. First, the vector $S$, which contains the ordered value of all the interstar angles, is computed. Second, the star pairs IDs are stored in two vectors labeled $I$ and $J$. Finally, the k-vector $K$ is gathered, where its $k$th element contains the number of elements of vector $S$ less than $\cos \bar{\gamma} = a_1 k + a_0$. The constants $a_1$ and $a_0$ are the coefficients describing the straight line that connects the first and the last element of $S$. The interested reader can consult [15] for further details about the method. In this work, to reduce the size of the catalog angles smaller than 35 deg are considered. Moreover, stars whose apparent magnitude is lower than 5.5 are considered for the generation of the invariant. The vectors $K$, $I$, and $J$, and the parameters $a_0$ and $a_1$ are stored on board and are used during onboard star identification. During the operational phase, star identification is performed by finding a set of possible correspondences between the measured inter-star angles and the values contained inside $K$. At this point, the IDs of the admissible catalog star pairs are determined by looking into $I$ and $J$. Finally, to select the right stars-pair among the possible ones, the Reference-Star method [15] is adopted. Similar performances in this last step can be achieved by exploiting the angle pivoting algorithm [27]. When the observed star asterism is recognized, the stars identification algorithm gives as output the vector of the stars identifiers and the matrices $[{}^{N}s]$ and $[{}^{C}s]$, whose columns contain the stars LoS directions in the $N$ and $C$ reference frame, respectively. Moreover, a vector including the position of the spikes in the image is delivered as well.

The objects identified by the k-vector as spikes may be non-stellar objects, such as planets, asteroids, and cosmic rays that are not present in the onboard catalog, or stars that have not been recognized due to errors in the centroid extraction. Yet, when a great number of spikes is present in the image, the star asterisms may not be recognized by the algorithm. In this work, to reduce the number of scenarios in which this failure occurs, a heuristic approach is considered. As faint stars are generally not stored in the onboard catalog and as the centroids extraction depends on the thresholding procedure, when the attitude determination fails, the attitude determination procedure is iterated again by increasing the tuning parameter $T$ of the intensity threshold $I_{\text{thr}}$ in Eq. (6). One of the results of this approach consists of diminishing the number of bright objects in the image, which can ultimately lead to the removal of some spikes. The procedure is repeated until observed star asterisms are recognized or less than three stars are detected.

*3. Attitude Determination*

Eventually, the probe attitude is determined by solving Wahba's problem [22] between the stars LoS directions in the camera and inertial reference frame. The Wahba's problem solution is computed by the Singular Value Decomposition (SVD) method [28]:

$$[B] = \sum_{i=1}^{n} {}^C s_i {}^N s_i^\top \qquad (10)$$

where $n$ is the number of identified stars in the image considered for attitude determination, and ${}^C s_i$ and ${}^N s_i$ are the $i$th columns of $[{}^C s]$ and $[{}^N s]$, respectively. As the $[B]$ matrix is not orthonormal due to measurement errors, the closest orthonormal matrix $[A]$ can be computed by imposing the matrix eigenvalues. Thus:

$$[B] = [U][S][V]^\top \rightarrow [A] = [U][M][V]^\top; \qquad (11)$$

where $[M]$ is used to impose a right-handed reference frame, and it is defined as:

$$[M] = \begin{bmatrix} 1 & 0 & 0 \\ 0 & 1 & 0 \\ 0 & 0 & \det[U]\det[V] \end{bmatrix} \qquad (12)$$

In this work, the robustness of the solution to Wahba's problem is increased thanks to the adoption of a RANdom SAmple Consensus (RANSAC) procedure [1]. The RANSAC algorithm of Fischler and Bolls [29] is a general and robust iterative method able to estimate the parameters of a mathematical model from a set of input data with a large proportion of outliers [16]. It can be also seen as an outliers detection and rejection method.

Here, the RANSAC algorithm aims to detect the bright objects that have been misidentified by the star identification step, which can thus lead to a wrong attitude determination. The star identification procedure can result in a misidentification when a non-stellar object is identified as a star or when a star is labeled with a wrong star identifier. To detect the outliers the attitude of the spacecraft is adopted as the mathematical model for the data fitting. The attitude is estimated $n_R$-times by selecting randomly every time a group of 3 identified stars. The minimum set of stars needed for attitude determination is chosen to increase the probability of having a group made of different stars at each time. Thus, the estimated $n_R$ spacecraft orientations are compared to identify the best model, which is then adopted for the data fitting. The stars not respecting the best model are considered outliers and are labeled as spikes. In detail, the RANSAC procedure can be summarized as follows:

1) A number of samples $n_R$ is selected.
2) For each $i$th sample with $i \in [1, n_R]$, a group of 3 stars is randomly selected within the identified stars.

3) For each group, Wahba's problem is solved, and the spacecraft rotation principal axis $e_i$ is defined. As said before, the attitude of the spacecraft is adopted as the mathematical model for the data fitting, and the rotation principal axis $e$ is the chosen attitude representation.

4) To each rotation principal axis $e_i$, a score is assigned dependent on the number of vectors $e_j$ ( related to the $j$th sample with $j \in [1, n_R]$ and $j \neq i$) that are within a threshold angle $t$ of $e_i$. The set of vectors that satisfy this requirement is called the consensus set of $e_i$. The size of the consensus set associated with $e_i$ can be determined as the score of $e_i$.

5) The vector $e_i$ characterized by the largest consensus set, thus, by the highest score, is selected as the best model.

6) The best model is then exploited for the data fitting: Only the stars that generate a subset related to a principal rotation vector contained inside the consensus set of the best model are considered inliers. The remaining stars (outliers) are identified as spikes. If two or more consensus sets are characterized by the same dimension, the best vector is chosen arbitrarily among the vectors $e_i$ related to these consensus sets. The probe attitude is, thus, redetermined by considering only the inlier stars.

A graphical representation of this process is depicted in Fig. 3. In this example, $n_R = 4$ for the sake of a clear graphical representation. For each of the four samples, a group of three stars $s_i$ is selected among the ones identified by the star identification step. Then, the associated principal rotation vector is computed for each group, and a score is assigned to each vector. In this case, a score of 2, 1, 1, 0 is assigned to $e_2$, $e_1$, $e_3$, and $e_4$ directions, respectively. The principal rotation vector $e_2$ has the highest score and, thus, is chosen as the best mathematical model for the data fitting. Since $e_3$ and $e_1$ vectors lie inside the consensus set of $e_2$, all the star subsets that are adopted to generate these three vectors are considered inliers. Whereas, the remaining stars are identified as outliers and, thus, labeled as spikes.

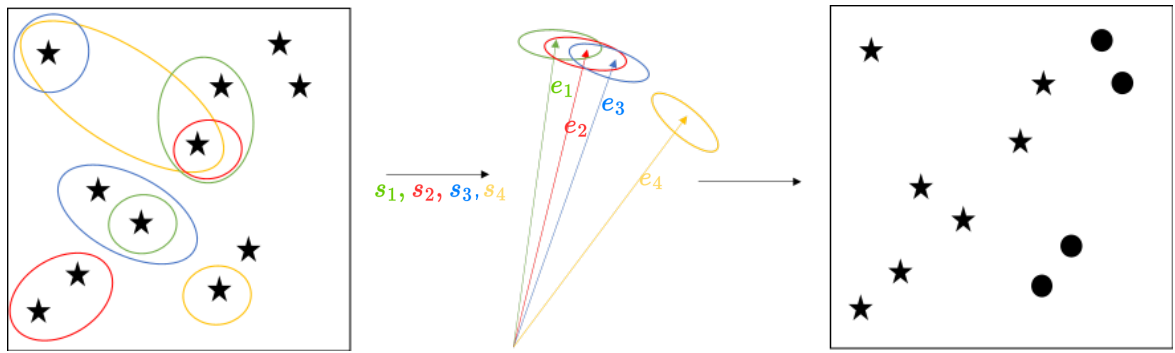

**Fig. 3   Graphical Representation of the RANSAC Algorithm**

## B. Beacon Detection

In this section, the second step of the proposed IP pipeline is presented. It starts with the detection of the beacon in the image and ends with the extraction of its associated position projection.

*1. Beacon Identification*

The beacon identification in the image is performed by computing the statistical momenta associated with the beacon projection, i.e., the expected beacon position projection and its covariance matrix, which define the $3\sigma$ Gaussian probability to find a beacon in that portion of the image. Thus, the following computation is carried out.
Once the attitude matrix $[A]$ is determined, and by assuming the beacon ephemerides and the probe position are known with a certain accuracy on each Cartesian component, the expected beacon position projection $^{\mathbb{C}}\boldsymbol{R}_{bc_0}$ is computed (Eqs. (4) and (5)). If the beacon $^{\mathbb{C}}\boldsymbol{R}_{bc_0}$ falls inside the image boundaries, its covariance matrix, which depends on the spacecraft pose and the beacon position uncertainties, can be defined. Once the image plane portion with the highest probability, here considered $3\sigma$, has been identified, this information is used to recognize the beacon in the image. If one or more spikes are contained in the $3\sigma$ uncertainty ellipse, the spike closest to the expected beacon position projection is labeled as its correct position projection. The closest one is selected because, from a statistical point of view, it is the one with the highest probability of being the projected beacon.
The covariance matrix of the beacon position projection $[P]$ due to the spacecraft pose and beacon position uncertainty is computed as

$$[P] = [F][S][F]^\top \tag{13}$$

where $[F]$ is the Jacobian matrix of the mapping linking the beacon position projection $^{\mathbb{C}}\boldsymbol{R}_{bc}$ with the spacecraft pose and the beacon position, and $[S]$ is the uncertainty covariance matrix of the probe pose and beacon position. To evaluate $[F]$, the variation of $^{\mathbb{C}}\boldsymbol{R}_{bc}$ with respect to the variation of the spacecraft pose and the beacon position has to be computed. To simplify the calculus, the quaternions $\boldsymbol{q} = (q_0, \boldsymbol{q}_v)^\top$, where $q_0$ is the scalar part and $\boldsymbol{q}_v$ is the vectorial part, are chosen to represent the probe attitude matrix. Eq. (14) gives the quaternion representation of the attitude matrix $[A]$ [30]

$$[A] = (q_0^2 - \boldsymbol{q}_v^\top \boldsymbol{q}_v)\,[\mathbb{I}_3] + 2\boldsymbol{q}_v \boldsymbol{q}_v^\top - 2q_0 [\boldsymbol{q}_v]^\wedge \tag{14}$$

Thus, the variation of $^{\mathbb{C}}\boldsymbol{R}_{\text{bc}}$ with respect to the the variation of the spacecraft pose, i.e., $[A(\boldsymbol{q}_{C/N})]$ and $^{N}\boldsymbol{r}$, and the beacon position $^{N}\boldsymbol{r}_{\text{bc}}$ can be defined as

$$\delta\,^{\mathbb{C}}\boldsymbol{R}_{\text{bc}} = \underbrace{\left[\frac{\partial\,^{\mathbb{C}}\boldsymbol{R}_{\text{bc}}}{\partial q_0}\quad \left[\frac{\partial\,^{\mathbb{C}}\boldsymbol{R}_{\text{bc}}}{\partial \boldsymbol{q}_v}\right]\quad \left[\frac{\partial\,^{\mathbb{C}}\boldsymbol{R}_{\text{bc}}}{\partial\,^{N}\boldsymbol{r}}\right]\quad \left[\frac{\partial\,^{\mathbb{C}}\boldsymbol{R}_{\text{bc}}}{\partial\,^{N}\boldsymbol{r}_{\text{bc}}}\right]\right]}_{[F]}\begin{pmatrix}\delta q_0 \\ \delta \boldsymbol{q}_v \\ \delta^{N}\boldsymbol{r} \\ \delta^{N}\boldsymbol{r}_{\text{bc}}\end{pmatrix} \quad (15)$$

The matrix $[F]$ has dimension $2 \times 10$ and, by computing the derivatives of Eq. (4), it is defined as

$$[F] = \begin{bmatrix} \dfrac{1}{^{\mathbb{C}}_{h}R_{\text{bc}_3}} & 0 & -\dfrac{^{\mathbb{C}}_{h}R_{\text{bc}_1}}{^{\mathbb{C}}_{h}R^2_{\text{bc}_3}} \\ 0 & \dfrac{1}{^{\mathbb{C}}_{h}R_{\text{bc}_3}} & -\dfrac{^{\mathbb{C}}_{h}R_{\text{bc}_2}}{^{\mathbb{C}}_{h}R^2_{\text{bc}_3}} \end{bmatrix}[K]\begin{bmatrix}\dfrac{\partial([A]\,^{N}\boldsymbol{\rho})}{\partial q_0} & \dfrac{\partial([A]\,^{N}\boldsymbol{\rho})}{\partial \boldsymbol{q}_v} & \dfrac{\partial([A]\,^{N}\boldsymbol{\rho})}{\partial\,^{N}\boldsymbol{r}} & \dfrac{\partial([A]\,^{N}\boldsymbol{\rho})}{\partial\,^{N}\boldsymbol{r}_{\text{bc}}}\end{bmatrix} \quad (16)$$

where the partial derivatives of $([A]\,^{N}\boldsymbol{\rho})$ with respect to the spacecraft pose and beacon position are

$$\frac{\partial([A]\,^{N}\boldsymbol{\rho})}{\partial q_0} = 2q_0\,^{N}\boldsymbol{\rho} - 2[\boldsymbol{q}_v]^{\wedge}\,^{N}\boldsymbol{\rho} \quad (17)$$

$$\frac{\partial([A]\,^{N}\boldsymbol{\rho})}{\partial \boldsymbol{q}_v} = -2\,^{N}\boldsymbol{\rho}\boldsymbol{q}_v^{\top} + 2\boldsymbol{q}_v^{\top}\,^{N}\boldsymbol{\rho}\,[\mathbb{I}_3] + 2\boldsymbol{q}_v\,^{N}\boldsymbol{\rho}^{\top} + 2q_0[^{N}\boldsymbol{\rho}]^{\wedge} \quad (18)$$

$$\frac{\partial([A]\,^{N}\boldsymbol{\rho})}{\partial^{N}\boldsymbol{r}} = -[A][\mathbb{I}_3] \quad (19)$$

$$\frac{\partial([A]\,^{N}\boldsymbol{\rho})}{\partial^{N}\boldsymbol{r}_{\text{bc}}} = [A][\mathbb{I}_3] \quad (20)$$

A change of attitude representation is performed to define $[S]$. Since the uncertainty of the probe orientation is more clearly identified through Euler's principal rotation theorem, the quaternion variation is linked to the one relative to the principal angle $\theta$ (pointing error) and principal axis $\boldsymbol{e}$. For small error angles, the following relations are verified [28]:

$$\delta q_0 = 0 \qquad\qquad \delta\boldsymbol{q}_v = \frac{1}{2}\delta(\theta\boldsymbol{e}) \quad (21)$$

Hence, $[S]$ can be described:

$$[S] = \text{diag}(\sigma^2_{q_0}, \sigma^2_{q_v}[\mathbb{I}_3], \sigma^2_{\text{r}}[\mathbb{I}_3], \sigma^2_{\text{r}_{\text{bc}}}[\mathbb{I}_3]) \quad (22)$$

where $\sigma_{\boldsymbol{r}}$ and $\sigma_{\boldsymbol{r}_{\text{bc}}}$ represent the standard deviation of the probe position and beacon position, respectively, and $\sigma^2_{q_0} = 0$ for the small-angles assumption. Note that the cross-correlations are ignored for simplicity, yet in a more integrated solution the pose could be coupled.

Once the covariance matrix of the beacon position projection is assessed, the associated $3\sigma$ uncertainty ellipse is computed. Let $\lambda_{\max}$ and $\lambda_{\min}$ be the largest and smallest eigenvalues of $[P]$, respectively, and $\boldsymbol{v}_{\max}$, $\boldsymbol{v}_{\min}$ their related eigenvectors. The characteristics of the $3\sigma$ covariance ellipse can be computed as:

$$a = \sqrt{11.8292\,\lambda_{\max}} \qquad b = \sqrt{11.8292\,\lambda_{\min}} \qquad \psi = \arctan\left(\frac{v_{\max_2}}{v_{\max_1}}\right) \tag{23}$$

where $a$ is the $3\sigma$ covariance ellipse semimajor axis, $b$ the $3\sigma$ covariance ellipse semiminor axis, $\psi$ the $3\sigma$ covariance ellipse orientation (angle of the largest eigenvector towards the image axis $\boldsymbol{C}_1$), and $v_{\max_2}$, $v_{\max_1}$ the eigenvector related to the maximum eigenvalue along $\boldsymbol{C}_2$ and $\boldsymbol{C}_1$ directions, respectively. Note that the value 11.8292 represents the inverse of the chi-square cdf with 2 degrees of freedom at the values in 0.9973 ($3\sigma$).

The equation of the uncertainty ellipse of the beacon position projection as a function of the angle $\theta$ is so derived

$$\begin{bmatrix} x \\ y \end{bmatrix} = \begin{bmatrix} a\cos\theta & b\sin\theta \end{bmatrix} \begin{bmatrix} \cos\psi & \sin\psi \\ -\sin\psi & \cos\psi \end{bmatrix} + {}^{\mathbb{C}}\boldsymbol{R}_{bc_0} \tag{24}$$

Eventually, the beacon is identified with the closest spike to the expected beacon position projection contained in the $3\sigma$ ellipse.

## IV. Algorithm Assessment

The methodology illustrated so far has a general application. Indeed, it can be adopted to detect any bright beacon, e.g., asteroids and planets, viable for deep-space VBN. Nevertheless, in this work, to test the performance of the IP pipeline, only planet ephemerides are included in the IP pipeline, and only planets are added to the image rendered with the sky-field simulator [18] as the proposed IP procedure is here tested in the framework of the EXTREMA project [31]. Since miniaturized cameras with limited performances are embarked onboard CubeSats, only planets are detectable by the optical sensor and can be detected by the algorithm [4].

Before entering the details of the numerical performances, it is worth providing a qualitative discussion about the possible off-nominal solutions that can be faced besides the estimation of the correct attitude and the correct identification of the planet. Indeed, despite the high rate of success of the IP pipeline, which depends mostly on the probe position uncertainty as shown in Sec. V, it is important to describe how the off-nominal conditions are verified and which heuristic approach can be adopted to avoid them. During the attitude determination and the planet detection, the algorithm can yield three results:

1) It can provide a solution, which is regarded as correct.

2) It can provide a solution, which is regarded as wrong.

3) It can not converge to a solution.

A more detailed description of these results is provided in the following subsections. First, the nominal and the off-nominal scenarios and the heuristic approaches adopted to prevent the off-nominal scenarios during the attitude determination step are described. Then, the same discussion is performed for the beacon detection step.

### A. Assessment of the Attitude Determination Procedure

The attitude determination step can yield three results:

I) The star identification algorithm converges to the correct solution

II) The star identification algorithm converges to a wrong solution, which is appointed when the pointing error is greater than 500 arcsec.

III) The star identification algorithm does not converge to a solution. This may be due to different reasons: there are less than three stars in the image; the background noise is too high to identify bright points, or too many spikes are present in the image (usually higher than the 25% [15]).

To decrease the number of scenarios in which the star asterism is not identified, the heuristic approach described in Sec. III.A.2 is implemented. Instead, the RANSAC algorithm described in Sec. III.A.3 is adopted to prevent stars' misidentification. Indeed, the RANSAC method may recognize the errors performed by the star identification step (non-stellar objects identified as stars or stars labeled with a wrong identifier) by appointing the misidentified object as a spike. In this way, misidentified stars are not considered by the pipeline when the attitude of the probe is determined. An analysis of the IP pipeline performances when the RANSAC is not applied is reported in [1].

### B. Assessment of the Beacon Detection Procedure

The IP pipeline applied to deep-space images generated by the DART Lab can yield three results: correct, wrong, or no identification of the planet. To assess the correctness of these performances, an external control algorithm, independent of the IP pipeline, is exploited to define whether the planet is visible from the camera or not. In other terms, if the camera can effectively detect it or not. The planet visibility is assessed by verifying that the planet must not only fall within the image borders but also have an intensity higher than the camera detectability level, which, in this work, it is set to 120 (out of 255). Since the camera detectability level is established on an arbitrary consideration, this choice may yield two extreme cases where: 1) the planet could be assessed as not visible by the control algorithm, but it could be detected from the IP pipeline anyway, or 2) the planet is assessed as visible by the control algorithm, even if it is highly faint, and the IP pipeline can not detect it (see off-nominal scenario 1.III.F)).

1) **Planet assessed visible in the camera image**. When the planet is visible in the image, the IP pipeline can yield three results:

1.I) The planet is correctly identified (see Fig. 4a).

1.II) The planet is wrongly identified, which occurs when the planet is associated with a wrong spike, thus, one not corresponding to its correct position. The celestial body is appointed to be wrongly identified when the distance between the real planet position projection (+) and the spike detected as the position projection of the planet (□) is greater than 5 px (see Fig. 4b).

1.III) The planet is not detected by the IP algorithm, although it is visible by the camera. In other terms, the planet is present and visible in the image, but the algorithm does not spot it.

Scenarios 1.II) and 1.III) are considered off-nominal scenarios of the IP pipeline. A situation that can lead to scenario 1.II) is:

1.II.A) The uncertainty of the spacecraft position is large. Thus, the expected beacon position projection is far from the real one. The uncertainty ellipse increases in size, which may lead to having as the closest spike to the expected beacon position projection not the correct one (see Fig. 4b). A numerical study on the probe position uncertainty is investigated in Sec. V.

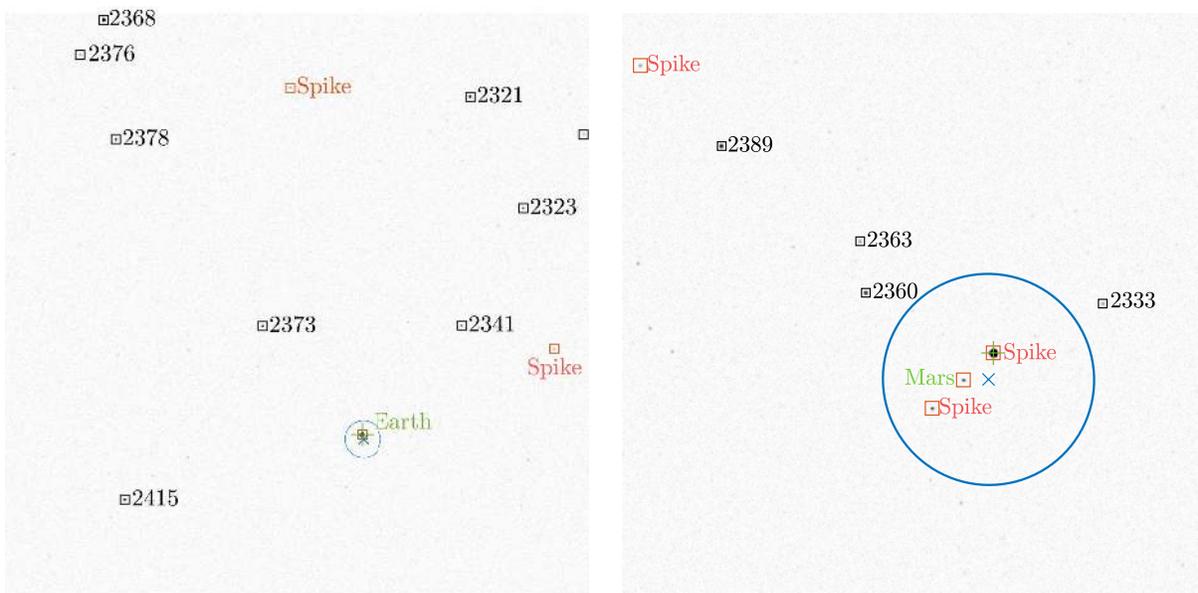

(a) Scenario 1.I). Planet correctly identified. The norm of the planet position projection error is 0.0965 px.

(b) Scenario 1.II). Planet wrongly identified. The planet position projection is associated to a wrong spike. The correct one would be where the green cross + is placed.

**Fig. 4** Representations of scenarios 1.I) and 1.II). + represents the real planet position projection, × represents the expected planet position projection, and □ the found spikes, respectively.

For what concerns scenario 1.III), six different events are identified as leading to it:

1.III.A) The IP pipeline mistakes the celestial body for a star. The expected position projection of the celestial body is detected inside the image, but no spikes are found inside the uncertainty ellipse associated. This scenario occurs since the spike that corresponds to the position projection of the beacon was wrongly identified as a star by the star identification algorithm. The approach applied to reduce the presence of this scenario consists of the RANSAC algorithm adoption, which assesses the misidentified stars as spikes. In this way, if a planet is identified first as a star by the star identification step, the RANSAC algorithm may recognize the error and relabel the object as a spike, among which the planet will be then searched.

1.III.B) The spike associated with the planet position projection falls outside the $3\sigma$ uncertainty ellipse, and thus it can not be detected by the algorithm (see Fig. 5a). Note that the $3\sigma$ represents a probability of 99.7%, thus, there are few possible scenarios in which the celestial body falls outside.

1.III.C) The attitude of the probe has been wrongly determined. Thus, planets are present and detectable in the image, but the pipeline is not able to recognize them (see Fig. 5b). This off-nominal scenario is assessed as a failure of the attitude determination algorithm ( see Scenario I)).

1.III.D) The celestial body is close to the image border. The expected planet position projection is outside the image, but the real one is inside. This scenario can be avoided by observing only the planets in the central part of the FoV.

1.III.E) The centroid associated with the planet position projection is not evaluated correctly. In the scenario shown in Fig. 5c two bright objects are contiguous. Thus, the centroiding algorithm finds only a centroid, instead of two, which is placed between the two objects. This centroid (the red square in the image) is outside the $3\sigma$ bounds of the error ellipse, so the planet is not detected.

1.III.F) The planet is considered visible from the camera, but no centroid is associated with it. Thus, it can not be detected by the IP pipeline (see Fig. 5d). This scenario is due to the parameters chosen for the thresholding procedure (Eq. 6) in the IP pipeline. This off-nominal scenario can be avoided during the operational phase by selecting only the brightest planets in the image.

2) **Planet assessed not visible in the camera image.** Three algorithm behaviors are identified when the planet is not visible in the camera image:

2.I) The beacon is not detected by the IP algorithm as its expected position projection is not in the image.

2.II) The beacon is not detected by the IP algorithm. The expected position projection of the planet is in the image, but no centroids are associated with it.

2.III) The beacon is detected by the IP pipeline, although there are no visible planets.

Only scenario 2.III) is considered off-nominal for the IP pipeline. For the identification of this event, the same method adopted to recognize the second scenario is applied. A scene that may lead to the scenario 2.III) is:

2.III.A) When the expected planet position projection is close to the image border, this one may fall inside the image, even though the spike related to the planet position projection is not contained in it. The failure occurs because, at this point, the planet is associated with a wrong spike present in the image.

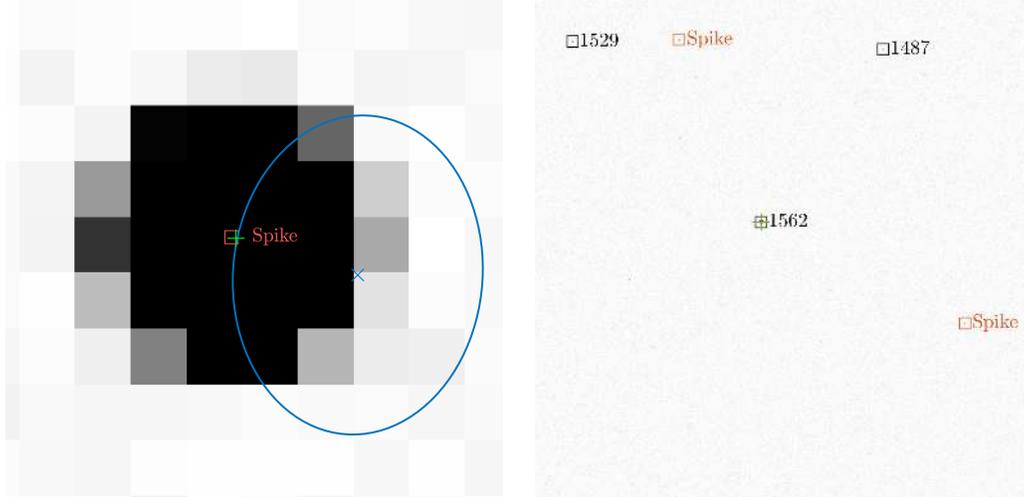

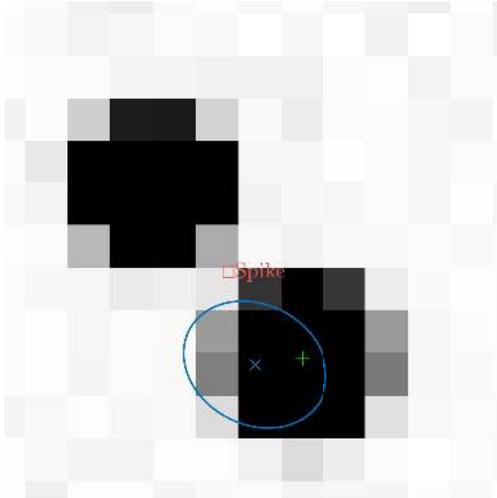 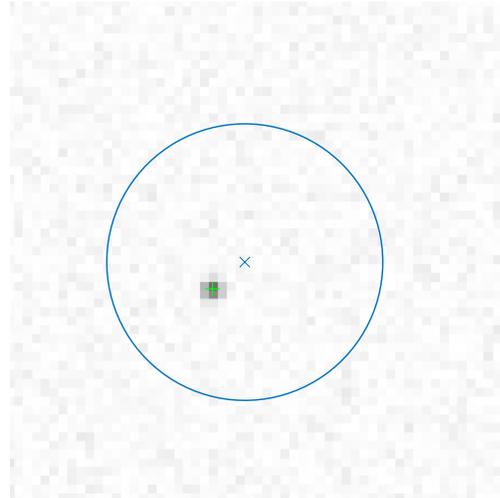

(a) Scenario 1.III.B).
The spike related to the beacon projection falls outside the $3\sigma$ ellipse.

(b) Scenario 1.III.C). The attitude is wrongly determined (err ang = $10^5$ arcsec). The planet is mistaken for the star whose ID is 1562.

(c) Scenario 1.III.E). The centroiding algorithm resolves only one centroid, instead of two, which is placed between the two objects.

(d) Scenario 1.III.F). The planet is so faint that the centroiding algorithm does not associated any spikes to it.

**Fig. 5  Representations of scenarios 1.I) and 1.II). + represents the real planet position projection, × represents the expected planet position projection, and □ the found spikes, respectively.**

## V. Results

A quantitative discussion about the performances of the IP pipeline is presented in this section.

### A. Simulation Settings

A Monte Carlo campaign is carried out to assess the performances of the developed algorithm. The extraction of the beacon position projection is run for 1031 scenarios, wherein at least one planet is present, out of the 3000 scenarios analyzed. In each scenario, the position of the spacecraft is selected by randomly sampling a Gaussian distribution with $\sigma_x = \sigma_y = 3$AU and $\sigma_z = 0.07$AU and centered in the origin of $\mathcal{N}$. The $z$-component of the probe position is chosen in a narrower interval as the spacecraft is supposed to lie close to the ecliptic plane. Similarly, the orientation of the probe is assigned by randomly sampling a normal distribution $\alpha$, $\delta$, and $\phi$ in the $3\sigma$ intervals $[0, 2\pi]$, $[-0.6, 0.6]$, and $[0, 2\pi]$, respectively. The declination $\delta$ is chosen in a narrower interval as planets are distributed close to the ecliptic plane. Once the probe pose is sampled, the sky-field image is generated by exploiting an improved version of the DART Lab sky simulator presented in [18]. The onboard camera characteristics are reported in Table 1, where F is the f-number, $Q_e \times T_{lens}$ is the quantum efficiency $\times$ lens transmission, SEA is the Solar Exclusion Angle, and $\sigma_d$ is the defocus level.

Table 1   Onboard camera characteristics.

| FoV [deg] | Image size [px] | f [mm] | F [-] | T [ms] | $Q_e \times T_{lens}$ | SEA [deg] | $\sigma_d$ [px] |
|---|---|---|---|---|---|---|---|
| 20 | 1024 × 1024 | 40 | 2.2 | 400 | 0.49 | 20 | 0.9 |

Moreover, although the DART Lab sky-field simulator includes the possibility to simulate the impact of cosmic rays hitting the camera detector [18], this capability is not exploited in this work as the detection of the saturated pixels can be performed outside the proposed IP pipeline. Indeed, cosmic rays are usually easy to identify for under-sampled images because they hit only single pixels *.

The settings adopted for the first section of the IP pipeline are listed in Table 2, where $T$ is the tuning parameter of Eq. 6, $\epsilon$ is the k-vector range error, $m_{lim}$ is the apparent magnitude threshold of the cataloged stars, $n_R$ is the RANSAC samples, and $t$ is the RANSAC threshold. Whereas, $\sigma_{q_0} = 0$ since $\delta q_0 = 0$ (Eq. 22), $\sigma_{\boldsymbol{q}_v}$ is set equal to $10^{-4}$ as results of a statistical analysis conducted on the error obtained in the attitude determination, and the planet position uncertainty $\sigma_{\boldsymbol{r}_{bc}}$ is assumed equal to zero because of the high accuracy with which the planets ephemeris are known. Instead, when other objects are observed to navigate, e.g., asteroids [5] or debris [32] (not in a deep-space application), their associated position uncertainty needs to be taken into account.

Table 2   Setup for the Attitude Determination algorithm.

| $T$ [-] | $\epsilon$ [arcsec] | $m_{lim}$ [-] | $n_R$ [-] | $t$ [arcsec] |
|---|---|---|---|---|
| 20 | 7 | 5.5 | 20 | 15 |

## B. Numerical Results

This section presents the numerical performances of the IP pipeline. The performance indexes adopted for the discussion are the angular error for the attitude determination and the beacon projection error for the planet detection step. A sensitivity analysis is performed to study the robustness of the IP pipeline to the initial uncertainty of the probe

---
*https://www.eso.org/~ohainaut/ccd/CCD_artifacts.html

position $\sigma_r$ when the latter is set to $10^4$, $10^5$, $10^6$, and $10^7$ km, respectively. Note that these values are chosen following a conservative approach and only with the goal of assessing the robustness of the IP pipeline. Indeed, in deep space, the probe initial position is usually known with an accuracy better than $10^4$ km.

The performances of the IP pipeline are shown in Table 3.

Table 3   Algorithm Performances

| $\sigma_r$ [km] | $\sigma_{ErrRot}$ [arcsec] | % Wrong Attitude Determination (out of 1031 cases) | % No Attitude Determination (of 1031 cases) | % Wrong Beacon Detection (of 962 cases) | % Wrong Beacon Detection with Right Attitude Determination (of 962 cases) |
|---|---|---|---|---|---|
| $10^4$ | 14.77 | 3.88 (40 cases) | 6.69 (69 cases) | 4.57 (44 cases) | 0.42 (4 cases) |
| $10^5$ | 15.18 | 3.88 (40 cases) | 6.69 (69 cases) | 4.68 (45 cases) | 0.52 (5 cases) |
| $10^6$ | 15.21 | 4.07 (42 cases) | 6.69 (69 cases) | 7.69 (74 cases) | 3.33 (32 cases) |
| $10^7$ | 15.48 | 4.07 (42 cases) | 6.69 (69 cases) | 29.83 (287 cases) | 25.47 (245 cases) |

The first part of the algorithm succeeds in the determination of the probe attitude in about 90% of the scenarios (of 1031) independently of the probe position uncertainty. The slight variation observed in the wrong attitude determination cases is due to the random behavior of the RANSAC algorithm. Whereas, the percentage of off-nominal scenarios during the planet identification, and, thus, during the extraction of its position projection, greatly depends on the probe position uncertainty. Indeed, when $\sigma_r$ increases, the expected planet position projection is further from the real position projection, and its uncertainty ellipse is bigger, which leads to a likelier planet misidentification (see Scenario 1.II.A)). Moreover, the percentage of off-nominal scenarios in planet detection also depends strictly on the success of the attitude determination. Indeed, when attitude determination provides a wrong solution, planet detection fails consequentially. In Table 3, the fifth column represents the total number of cases of no detection or wrong identification when the attitude determination converges to a solution. The total number of scenes in which attitude determination converges to a solution is 962. Instead, the last column represents the number of cases of no detection or wrong identification of the planet when the attitude determination converges to the correct solution. The failure percentage of the beacon detection procedure when the probe attitude is correctly determined is lower than 1 % with a probe position uncertainty up to $10^5$ km. Thus, if a more robust attitude determination procedure is adopted, the total percentage of failure in the beacon detection becomes remarkably lower.

The Gaussian distribution of the planet position projection errors for $\sigma_r = 10^4$ km, $\sigma_r = 10^5$ km, $\sigma_r = 10^6$ km, and $\sigma_r = 10^7$ km is shown in Figs. 6a, 6b, 6c, and 6d, respectively. The color bar represents the number of samples lying in each grid interval. When the probe position uncertainty increases, the scenarios where the beacon projection error is over 0.3 pix seem to be filtered out. Indeed, in these cases, the IP algorithm may select a different, wrong, spike as the expected position projection becomes far from the real one (see Scenario 1.II.A)). As a result, the error norm becomes greater than 5 pix, and it is, thus, regarded as a failure of the IP procedure and not represented in pdf distributions.

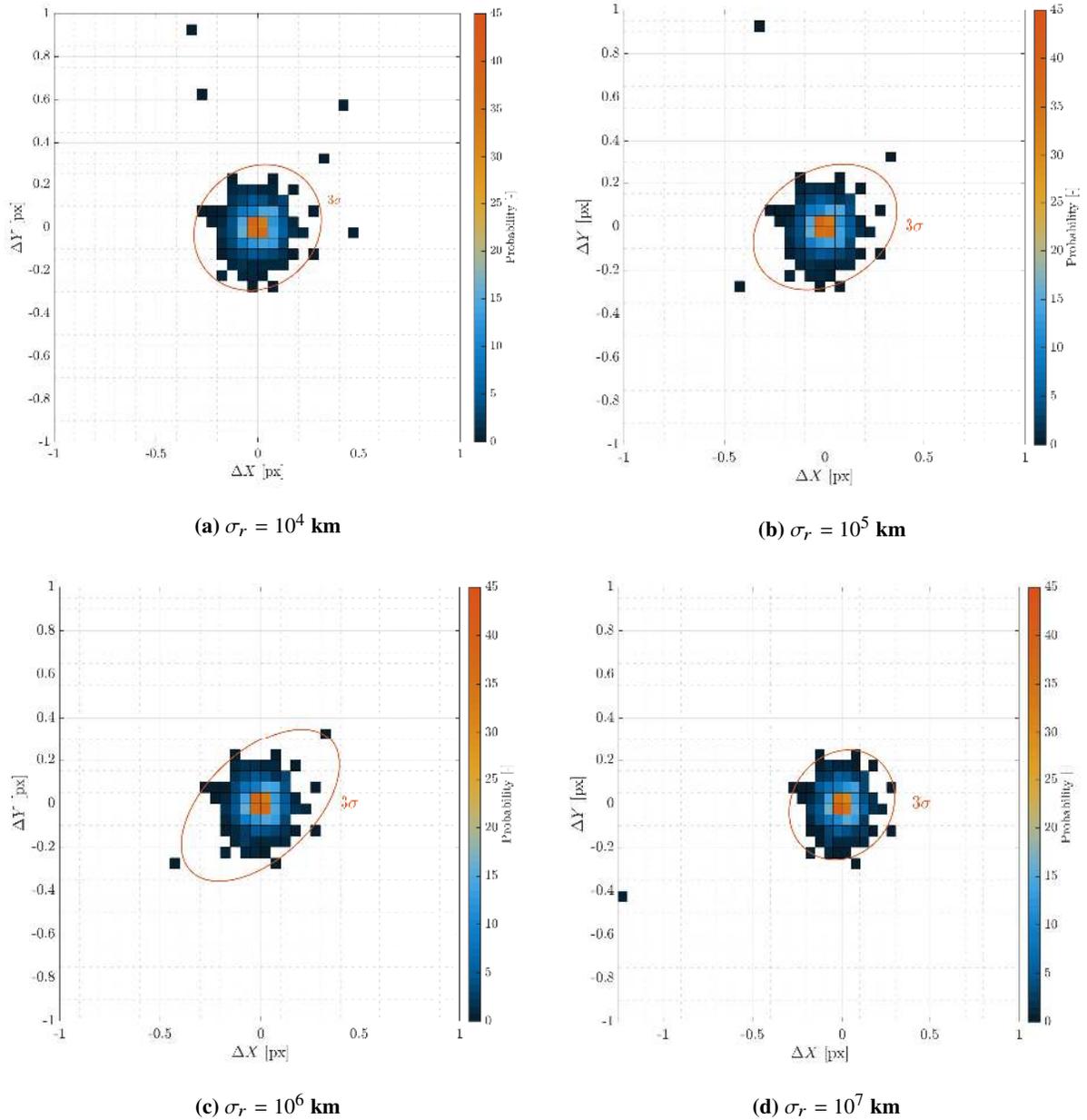

**Fig. 6   Pdf distribution of the planet position projection errors with $3\sigma$ bounds.**

The error ellipses in Fig. 6 are described by the mean and covariance values reported in Table 4. The determinant of the covariance matrix is a representation of the size of the area of the ellipse. Note that the planet position projection is detected with a sub-pixel $3\sigma$ accuracy for all the four values of $\sigma_r$. In other terms, the error on the estimated planet position projection is not dependent on the probe position uncertainty but only on the attitude determination and centroids computation errors. The four covariance matrices are characterized by a similar determinant, which is proportional to the area of the ellipse. This feature is one of the advantages of the proposed pipeline for the planet detection in deep-space images. Fig. 7 shows an example scenario where the error is over one pixel. In this case, two bright objects, where one of these is a planet, are overlapped. The centroiding algorithm finds only a centroid, shown with a red square, for this configuration shifted by more than one pixel from the planet real position projection shown with a green cross. Even in this challenging scenario, the IP pipeline can recognize the planet, but the detected planet position projection is affected by a greater error due to this unfortunate geometrical configuration.

**Table 4   Mean and covariance of the planet position projection errors when the probe position uncertainty is known with an accuracy of $10^4$, $10^5$, $10^6$, and $10^7$ km.**

| $\sigma_r$ [km] | $10^4$ | $10^5$ | $10^6$ | $10^7$ |
|---|---|---|---|---|
| $[P_{\text{err}}]$ [px$^2$] | $\begin{bmatrix} 0.008 & 0.001 \\ 0.001 & 0.007 \end{bmatrix}$ | $\begin{bmatrix} 0.011 & 0.002 \\ 0.002 & 0.007 \end{bmatrix}$ | $\begin{bmatrix} 0.013 & 0.006 \\ 0.006 & 0.011 \end{bmatrix}$ | $\begin{bmatrix} 0.007 & 0.001 \\ 0.001 & 0.005 \end{bmatrix}$ |
| $\mu_{\text{err}}$ [px] | [0.0014;-0.0003] | [0.0001;-0.0034] | [-0.0005; -0.0045] | [-0.0005;-0.0041] |
| det([P]) [px$^4$] | 5.9e-05 | 7.05e-05 | 9.94e-05 | 7.05e-05 |

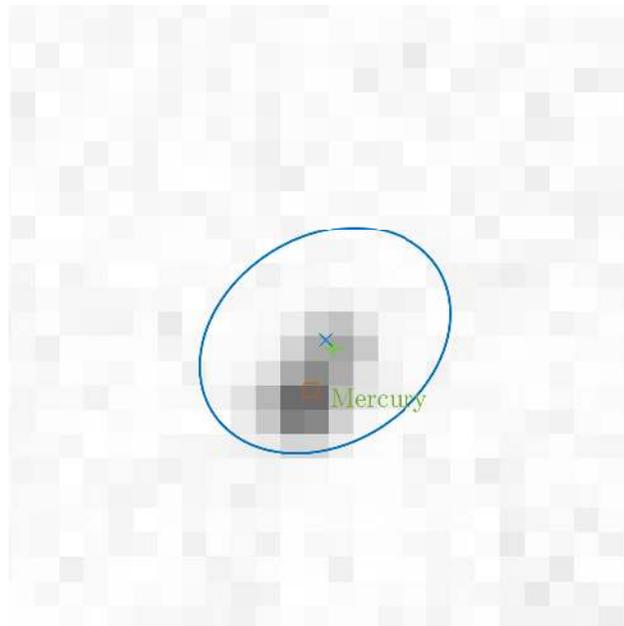

**Fig. 7   An example scenario where the planet position projection is affected by an error of more than one pixel.**

## VI. Conclusion

This work proposes a novel and robust beacon detection algorithm for deep-space vision-based navigation. The extracted planet projection in digital images enables the beacon LoS extraction, which paves the way for deep-space navigation by exploiting celestial triangulation.

The algorithm succeeds in the detection of the planet position projection in at least 95 % of the 962 tested scenarios when the probe position uncertainty is up to $10^5$ km, and a solution for the attitude determination problem is found. Since the rate of failure of the beacon detection is strictly connected to the success of the star identification procedure, the former can be reduced if a more robust procedure for the star asterism identification is exploited, as proposed in Mortari et al. [33] or in Cole and Crassidis [27], whose success rate is 95.8 % and 95 %, respectively. Indeed, the failure percentage of the beacon detection procedure when the probe attitude is correctly determined is lower than 1% with probe position uncertainty up to $10^5$ km.

Moreover, in this work, the IP algorithm is used to detect only planets, but the proposed pipeline is thought to be applied to other non-stellar objects. For example, by including the asteroids ephemerides and the uncertainty of their position projection, the IP algorithm can be adapted to detect asteroids in the image and use them as beacons to navigate. In addition, it can be also specialized to detect Anthropogenic Space Objects for Earth-orbiting satellite navigation [32]. Moreover, it has been noticed that the size of the planet error ellipse increases as the distance probe–planet decreases. Indeed, the angle between the real planet LoS direction and the expected one is larger when the planet is close to the spacecraft. Thus, an higher uncertainty is associated with the closest planets, which means that a misidentification is likelier to occur for them. On the contrary, [34] proves that the vicinity of the planets to the probe is a valuable feature for increasing the state estimation accuracy. Thus, a trade-off between these two features needs to be performed to select the best pair of beacons to track, with which observations the state estimator is fed.

Future analysis should test the performances of the IP pipeline during hardware-in-the-loop simulations. In this context, a camera acquires a star-field image, rendered on a high-resolution screen, and gives the associated matrix of digital counts to the IP algorithm [35]. In addition, future works should focus on the integration of the proposed IP pipeline with orbit determination filters to complete the navigation cycle [4].

## Acknowledgments

This research is part of EXTREMA, a project that has received funding from the European Research Council (ERC) under the European Union's Horizon 2020 research and innovation programme (Grant Agreement No. 864697).